# Quality assurance of organs-at-risk delineation in radiotherapy

Yihao Zhao[1], Cuiyun Yuan[2], Ying Liang[2], Yang Li[2], Chunxia Li[3], Man Zhao[2], Jun Hu[1], Wei Liu[4], Chenbin Liu[2]*

1. School of Electronics and Communication Engineering, Sun Yat-sen University, Shenzhen, Guangdong, China
2. National Cancer Center/National Clinical Research Center for Cancer/Cancer Hospital & Shenzhen Hospital, Chinese Academy of Medical Sciences and Peking Union Medical College, Shenzhen, Guangdong, China
3. Faculty of Health Sciences, University of Macau, Macau, China
4. Department of Radiation Oncology, Mayo Clinic Arizona, Phoenix, AZ, USA

Corresponding author: Chenbin Liu, chenbin.liu@gmail.com

**Abstract:**

**Purpose:** The delineation of tumor target and organs-at-risk (OARs) is critical in the radiotherapy treatment planning. It is also tedious, time-consuming, and prone to subjective experiences. Automatic segmentation can be used to reduce the physician's workload and improve the consistency. However, the quality assurance of the automatic segmentation is still an unmet need in clinical practice.

**Materials and Methods:** The patient data used in our study was a standardized dataset from AAPM Thoracic Auto-Segmentation Challenge. The OARs included were left and right lungs, heart, esophagus, and spinal cord. Two groups of OARs were generated, the benchmark dataset manually contoured by experienced physicians and the test dataset automatically created using a software "AccuContour^TM" (Manteia Medical Technologies Co. Ltd., Xiamen, China). A resnet-152 network was performed as feature extractor, and one-class support vector classifier was used to determine the 'high' or 'low' quality. We evaluate the model's performance with balanced accuracy, F-score, sensitivity, specificity and the area under the receiving operator characteristic curve (AUC). We randomly generated contour errors to assess the generalization of our method, explored the detection limit, and evaluated the correlations between detection limit and various metrics such as volume, Dice similarity coefficient (DSC), 95% Hausdorff distance (HD95), and mean surface distance (MSD).

**Results:** The proposed one-class classifier outperformed in metrics such as balanced accuracy, AUC, and others. The proposed model exhibited higher balanced accuracy compared to the CNN model (Esophagus: 0.96 vs. 0.92; heart: 0.98 vs. 0.95; left lung: 0.99 vs. 0.98; right lung: 0.99 vs. 0.98; spinal cord: 0.96 vs. 0.91). The AUC of the proposed model exceeded that of the CNN model (Esophagus: 0.96 vs. 0.95; heart: 0.97 vs. 0.95; left lung: 0.97 vs. 0.93; right lung: 0.97 vs. 0.94; spinal cord: 0.95 vs. 0.91). The proposed method showed significant improvement over binary classifiers in handling various types of errors. The relationships between the detection limit and multiple metrics of the organ indicate that our method is highly interpretable.

**Conclusion：** Our proposed model, which introduces residual network and attention mechanism in the one-class classification framework, was able to detect the various types of OAR contour errors with high accuracy. The proposed method can significantly reduce the burden of physician review for contour delineation.

# 1. Introduction

Contour delineation is a crucial step in radiotherapy treatment planning. It requires careful identification and outlining of targets and organs-at-risk to guide plan development. [1-3] However, manual contouring is tedious and prone to subjective experience, workload fatigue, and other influences. Accurate target and organs-at-risk (OARs) delineation is fundamental as subsequent planning and treatment. While the involvement of experienced physicians for reviewing and verifying delineations is important for quality assurance, the process remains time-consuming in clinical practice. It is also important to acknowledge that inter-observer and intra-observer variability may arise among physicians.[4] Automating or assisting contouring could help address these issues to improve consistency, efficiency and treatment outcomes.

According to the extent of user interaction, segmentation techniques could be categorized into manual, automatic, and semi-automatic methods. In automatic contouring methods, there are mainly two types: atlas-based and deep learning methods. Atlas-based methods created atlas from previously annotated dataset, deformed the atlas template(s) to the target images, and generated the target anatomical structures.[5,6] Deep learning methods applied multiple level of filters and max-pooling processes to extract image features (encoder), and inflated the encoder's output into a segmentation mask using convolution and up-sampling (decoder). Deep learning has demonstrated superior performance in image segmentation, including U-Net [7], 3D U-Net [8], V-Net[9], Seg-Net [10], DeepMedic[11], DeepLab[12], VoxResNet[13] and Mask RCNN [14]. However, deep learning models learn to recognize patterns in the data they are trained on, and if the new data does not follow the same patterns, the model will not be able to make accurate segmentation.[15] The performance of automatic delineation model may be limited by variations of clinical guidelines utilized and image acquisition protocols which may not be consistent with practice pattern.[16]

To mitigate the substantial time and resource costs associated with reviewing the delineations for each case, there is a need to develop an efficient and labor-effective method for identifying low-quality segmentations. Conventional practice employed an independent test dataset and diverse metrics, such as Dice similarity coefficient (DSC), Hausdorff distance ($HD_{95}$), and mean surface distance (MSD), to evaluate the quality of auto-segmentation. [17] These metrics provide a comprehensive assessment of the accuracy and robustness of the auto-segmentation algorithm. Previous researchers have also designed methods for automated quality control,[18-27] including feature-based methods, [20,22,28,29], and deep learning based methods.[19,21,25,26] Feature-based methods employed in quality prediction involved the extraction of features, such as DSC, $HD_{95}$, texture features, contour volume, surface area, and orientation to predict contour quality. These methods required hand-crafted features designed by clinical experts, had limited representation of complex patterns, and was prone to overfitting. In recent years, convolutional neural network (CNN) became a powerful category of deep neural networks that have been widely used in image analysis. It is particularly well-suited for tasks such as image classification, object detection, and semantic segmentation [30,31]. Rhee et al. [19] presented a CNN-based autocontouring tool for 16 head and neck normal structures. The tool was able to detect errors in contours generated by a clinically validated multiatlas-based autocontouring system. Chen et al. [26] developed a novel deep learning approach for the automated contouring quality assessment. Their method utilized DSC value as a metric to categorize contouring quality into three levels: good, medium, and bad. Duan et al. [25] proposed an improved contouring quality assurance system based on multiple geometric agreement metrics. The model was trained to identify potential contouring errors using 27 geometric features, and feature selection was performed to optimize the model.

Prior research efforts have focused on developing either standalone autocontouring systems or binary classifiers for contour quality assessment. While standalone autocontouring systems offered the potential for improved contouring accuracy, their clinical adoption may be limited if not approved for contour use but for quality evaluation. Binary classifiers, on the other hand, excelled at identifying specific contouring errors with high precision. However, their ability to handle complex contouring scenarios could be limited. One-class classifier was able to train a model in the absence of counter-examples. [33-36] It is widely used in abnormal detection. As far as we know, one-class classifier was not utilized for quality assurance in organ-at-risk (OAR) delineation. By incorporating a one-class classifier, we can address the challenge of identifying various errors and enhance the compatibility of the quality assurance process. Previous studies usually applied the whole CT image as the input of the quality assurance system. CNN may extract a substantial amount of irrelevant information out of the OARs. Inspired by He's work, we used the contoured regions similar to mask R-CNN to improve the identification accuracy. [14]

In this study, we proposed a one class support vector machine (OC-SVM) based method that initially extracted the images using OAR contours, then utilized ResNet-152 for image feature extraction, and finally employed a one-class classifier to predict the quality of OAR delineation. The extracted images using masks effectively addresses the issue of correlating delineations with CT images. ResNet-152 has proven to exhibit excellent performance in feature extraction. The one-class classifier was able to address the challenges of identifying various errors and the scarcity of error samples. We employed various methods to evaluate the performance of the proposed model, and explored its detection limit.

## 2. Materials and Methods

As shown in Figure 1, this section provides an overview of several key components in our process, including data acquisition, contour evaluation, data preprocessing, feature extraction, machine learning model, and evaluation metrics.

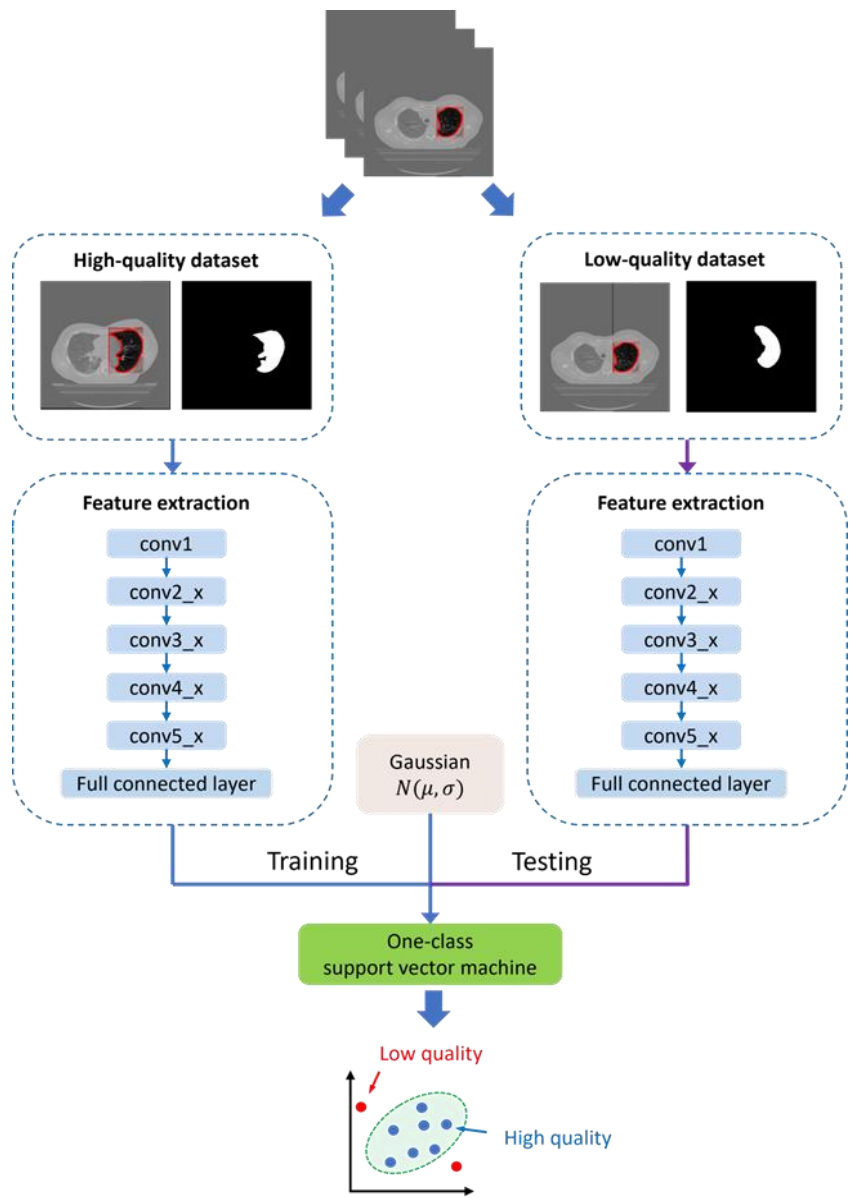

**Figure 1**. The whole workflow of the proposed quality assurance model.

### 2.1. Data Acquisition

The patient data utilized in our study comprised CT images and manually delineated contours from the AAPM Thoracic CT Segmentation Challenge competition, encompassing a total of 60 cases [38]. The CT images were reconstructed to encompass the entire thoracic region, with the number of slices ranging from 103 to 279. These CT scans had a consistent field of view of 50 cm and a reconstruction matrix size of 512 x

512. There was variation in slice spacing across institutions, ranging from 1 mm to 3 mm. The reported pixel size of the images ranged between 0.98 mm and 1.37 mm, with a median value of 0.98 mm.

The gold standard atlas for organ-at-risk (OAR) delineation in this study consisted of manual contours. The OARs included the left and right lungs, heart, esophagus, and spinal cord. These manual contours were created by expert clinicians following the contouring atlas guideline outlined in RTOG 1106 [39]. They served as a reference for accurately identifying and delineating these anatomical structures and were considered as the "ground truth" for comparison. To automatically generate contours, a commercially available contouring software, AccuContour™ (Manteia Medical Technologies Co. Ltd., Xiamen, China), was employed, utilizing deep learning techniques. The auto-generated contours were divided into two datasets based on quality: a high-quality contour dataset and a low-quality contour dataset. The objective of this study was to develop a quality assurance model capable of identifying low-quality contours. A total of 60 cases were included in this study, with 48 cases allocated to the training set and 12 cases to the test set. The selection of cases for both sets was performed randomly.

2.2. Contour Evaluation

We used Dice similarity coefficient (DSC) [40], the maximum Hausdorff distance ($HD_{95}$) [41], and mean surface distance (MSD) [42] to measure the contour quality. They were calculated by the follows:

$$DSC(GT, AGC) = \frac{2|GT \cap AGC|}{|GT|+|AGC|} \quad (1)$$

where GT is the ground truth and AGC is the automatically generated contours.

$$d(X \to Y) = \max_{x \in X} \min_{y \in Y} (d^{X \to Y}) \quad (2)$$

$$HD_{95}(GT, AGC) = max\,(d(GT \to AGC), d(AGC \to GT)) \quad (3)$$

where d is the one-sided Euclidean distance from point set X to point set Y. $HD_{95}$ is the longest bidirectional distance between the ground truth and automatically generated contours.

$$MSD(GT, AGC) = \frac{1}{N_{GT} + N_{AGC}} \left( \sum_{x \in GT} \min_{y \in AGC} \|x - S(AGC)\| + \sum_{y \in AGC} \min_{x \in GT} \|y - S(GT)\| \right) \quad (4)$$

where $N_{GT}$ and $N_{AGC}$ are the number of the pixels in the contour of ground truth and automatically generated contours respectively. $\|\cdot\|$ denotes the Euclidean distance. $S(\cdot)$ denote the point set of surface voxels.

Based on the aforementioned measurement, we established the criteria for contour quality. The high-quality contour was defined as one that met all the following requirements:

$$DSC > mean_{DSC} + \sigma_{DSC} \quad (5)$$

$$HD_{95} < mean_{HD95} - \sigma_{HD95} \quad (6)$$

$$MSD < mean_{MSD} - \sigma_{MSD} \quad (7)$$

In the equations (5)-(7), $mean_{DSC}$, $mean_{HD}$, and $mean_{MSD}$ are the average value of DSC, $HD_{95}$ and MSD calculated in the training set, respectively. $\sigma_{DSC}$, $\sigma_{HD95}$, and $\sigma_{MSD}$ are the standard deviation of DSC, $HD_{95}$ and MSD calculated in the training set, respectively. Contours that fail to meet any of the aforementioned requirements were labeled as low-quality contours. In our dataset, the proportion of low-quality contours in different organs ranges from 12% to 16%. Eight representative contours were shown in figure 2.

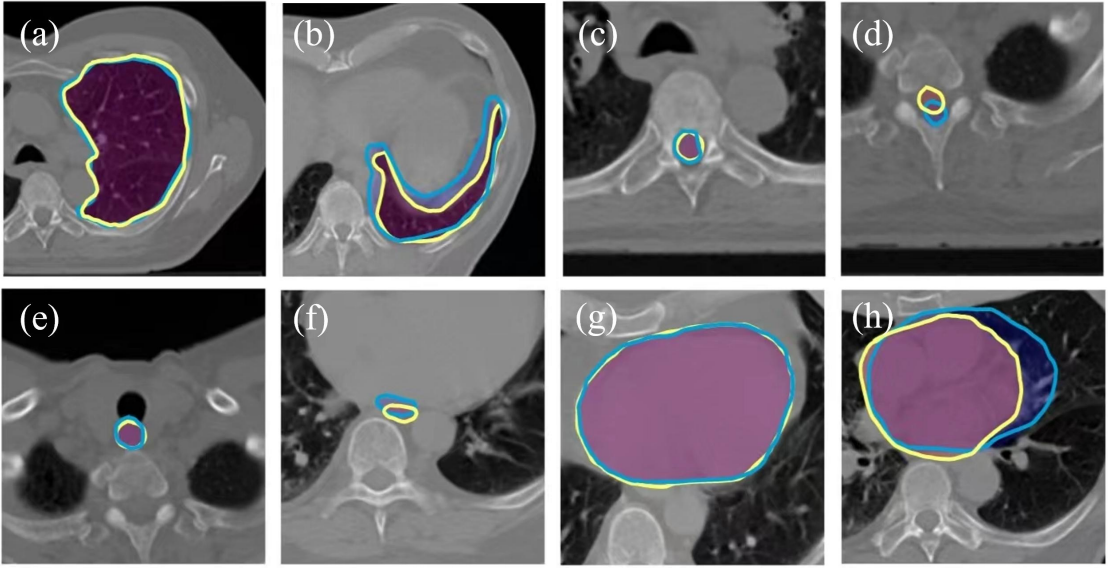

**Figure 2**. The representative contours created by expert clinicians and deep learning techniques. (a) high quality contour of left lung; (b) low quality contour of left lung; (c) high quality contour of spinal cord; (d) low quality contour of spinal cord; (e) high quality contour of esophagus; (f) low quality contour of esophagus; (g) high quality contour of heart; (h) low quality contour of heart. The yellow lines were contours generated by expert clinicians. The blue lines were contours created by deep learning techniques.

### 2.3. Data Preprocessing

To accommodate the variability in CT scan formats from different sources, we performed normalization by converting each scan into uint8 format, where pixel values ranged from 0 to 255. This normalization not only standardized the pixel value range, but also resulted in reduced training time and GPU memory usage. Subsequently, we converted both the manual and automatic generated contour into binary mask. By identifying non-zero pixels in the binary mask, we extracted corresponding regions of interest from the CT scans, which served as training images. To meet the input requirements of the ResNet-152 network [30], we resized the images to dimensions of $224 \times 224$.

### 2.4. Feature Extraction

We used Resnet-152 as the feature extractor of our model [30]. Its network architecture introduced the concept of residual learning, addressing the issues of vanishing and exploding gradients during the training of deep neural networks through skip connections and identity mappings across layers. ResNet-152 has gained renown for its exceptional depth and high performance, characterized by its 152 layers and proficiency in large-scale image recognition tasks. The key innovation of this network lies in the introduction of residual blocks, enabling the model to learn residual mappings. This meant the network could be optimized by learning the residual between the target mapping and the input and our model will have a better performance [30].

2.5. One-class Support Vector Machine

Due to the potential presence of various error types within the automatically generated contours and the predominant inclusion of high-quality contours in the training data, a binary classifier could encounter challenges associated with an imbalanced dataset. One-class classification algorithms, referred to as outlier or anomaly detection, is designed to identify instances that differ significantly from the majority class [34]. Its primary objective is to determine whether a given instance belongs to the target class or not, without explicit knowledge or training examples from other classes. In this study, one class support vector machine (OC-SVM) [34] was used. The advantage of OC-SVM is its capacity to detect diverse types of errors, even those that were not encountered during the training process. Furthermore, OC-SVM exhibits superior error detection capabilities compared to a binary classifier. The objective of the OC-SVM found a maximum margin hyperplane in feature space. It solved the following function:

$$\min_{\omega,b,\xi} \frac{1}{2}\|\omega\|_{F_k}^2 - \rho + \frac{1}{\nu n}\sum_{i=1}^{n} \xi_i \qquad (8)$$

where $\omega \in F_k$, $\rho$ is the distance from origin to hyperplane $\omega$. Nonnegative slack variables $\xi_i$ allow the margin to be soft, but violations $\xi_i$ get penalized. $\|\omega\|_{F_k}^2$ is a regularizer on the hyperplane $\omega$ and $\|\cdot\|_{F_k}^2$ is the norm induced by $\langle\cdot,\cdot\rangle_{F_k}$. In our study, OC-SVM was trained using high-quality contours and subsequently employed to identify low-quality contours. During the training of OC-SVM (Figure 1), we introduced zero-mean Gaussian noises as abnormal samples to the fully connected layer [39].

### 2.6. Evaluation

### 2.6.1 Prediction Evaluation

We used balanced accuracy, F score, sensitivity, specificity, and AUC to measure the performance of OC-SVM model. Balanced accuracy is a metric that helps mitigate sample size imbalances. It is defined as follow:

$$BA = \frac{1}{2}\left(\frac{TP}{TP+FN} + \frac{TN}{TN+FP}\right) \qquad (9)$$

where TP is the number of the samples with low-quality contour that are accurately identified by the model, FN represents the number of samples with high-quality contour that are inaccurately predicted, TN is the number of samples with high-quality contour that are correctly detected, and FP is the number of samples with contour errors that are incorrectly identified.

F-score measures the accuracy of a prediction model on a dataset. It is the harmonic mean of the precision and recall:

$$F\ score = \frac{2TP}{2TP+FN+FP} \qquad (10)$$

Sensitivity is the probability of a low-quality test result, given that the contour is truly low-quality, which evaluates the ability to identify low-quality contours. Specificity is the probability of a high-quality test result, given that the contour is truly high-quality, which reflects the ability to identify high-quality contours. They are defined as follows:

$$Sensitivity = \frac{TP}{TP+FN} \qquad (11)$$

$$Specificity = \frac{TN}{TN+FP} \qquad (12)$$

### 2.6.2 Extended Evaluation using Generated Errors

To expand the sample size and thoroughly evaluate the performance of OC-SVM, we additionally generated various types of errors derived from the automatically generated contours. The errors encompassed translation and resizing. To simulate a translation error, we started by randomly selecting a direction and subsequently applied a displacement to the contour using the direction and a designated distance. To create an enlargement error, a dilation kernel with a disk of radius 2 was employed to progressively enlarge the contour until the low-quality was achieved. The reduction error was generated using an erosion kernel, following a similar procedure as the enlargement error, but with the aim of reducing the contour size. These generated contours that did not meet the criteria (equation 6-8) mentioned above were labeled as low-quality as shown in Figure 3. For each high-quality contour, we generated three distinct types of low-quality contours, resulting in a total number of the generated low-quality contours of 2634.

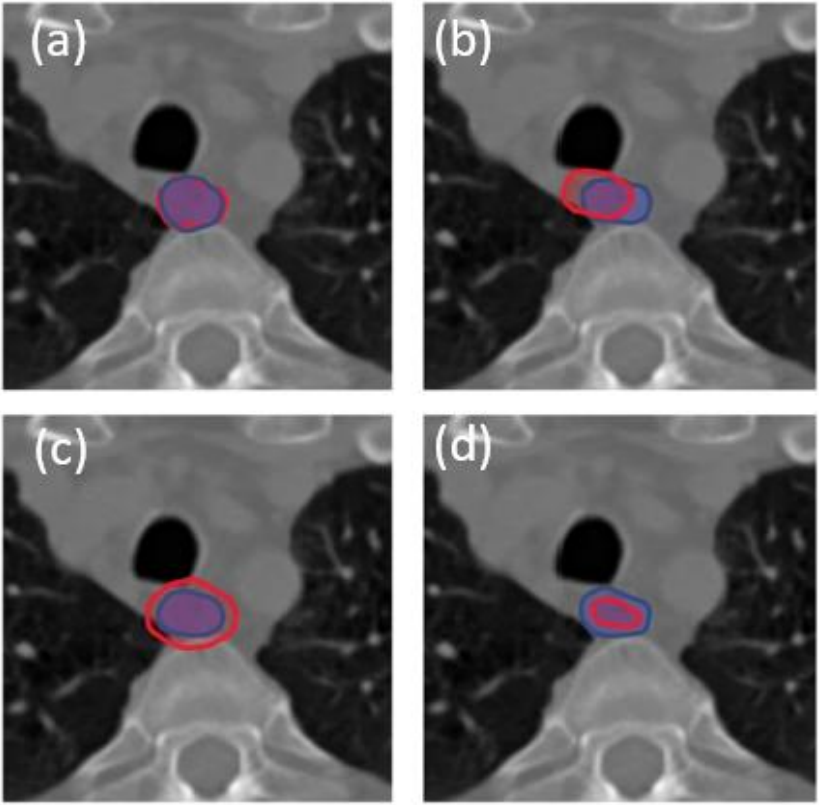

**Figure 3**. The contours with generated errors. (a) The automatically generated contour of the esophagus was labeled as high-quality, (b) the esophagus contour with translation error, (c) the esophagus contour with an enlargement error. (d) the esophagus contour with a shrinkage error

### 2.6.3 The detection limit

To explore the detection limit of the quality assurance model, we conducted a series of repeated experiments to determine the magnitude of translational errors at which the identification accuracy drops below 90%. In this experiment, we randomly perturbed the OAR contour by a given distance in a random direction, considering the perturbed contours as instances of low-quality contours. Subsequently, we evaluated the performance of the proposed model and CNN model on this set. We iteratively increased the specified distance until the accuracy dropped below 90%, which we identified as the detection limit.

To highlight the advantages of a one-class classifier in identifying unknown errors, we integrated the traditional CNN method [43] with an OAR mask. In our experiment, an NVIDIA GeForce RTX 3080 was used in the proposed quality assurance model (Figure 1). Due to the limitation of our GPU memory, the batch size was set to 32. We used Adam optimizer in the training of ResNet-152. The total epochs was set to 50. The training process took about 3 hours. Testing involved identifying 2000 contours within 5 minutes. On average, the prediction of quality assurance model for the contour on a single slice took 150 ms.

## 3. Results

### 3.1. Comparison of CNN model and the proposed model

We evaluated the performance of CNN model and our proposed quality assurance model on the test dataset. As shown in table 1, the proposed model exhibited higher balanced accuracy compared to the CNN model (Esophagus: 0.96 vs. 0.92; heart: 0.98 vs. 0.95; left lung: 0.99 vs. 0.98; right lung: 0.99 vs. 0.98; spinal cord: 0.96 vs. 0.91). The F-scores of the proposed model outperformed those of the CNN model (Esophagus: 0.97 vs. 0.93; heart: 0.98 vs. 0.96; left lung: 0.98 vs. 0.91; right lung: 0.98 vs. 0.92; spinal cord: 0.96 vs. 0.92). The sensitivity of the proposed model was superior to that of the CNN model (Esophagus: 0.96 vs. 0.93; heart: 0.97 vs. 0.96; left lung: 0.98 vs. 0.96; right lung: 0.98 vs. 0.97; spinal cord: 0.94 vs. 0.93). The specificity of the proposed model surpassed that of the CNN model (Esophagus: 0.98 vs. 0.91; heart: 1.00 vs. 0.94; left lung: 1.00 vs. 1.00; right lung: 1.00 vs. 1.00; spinal cord: 0.98 vs. 0.89). The AUC of the proposed model exceeded that of the CNN model (Esophagus: 0.96 vs. 0.95; heart: 0.97 vs. 0.95; left lung: 0.97 vs. 0.93; right lung: 0.97 vs. 0.94; spinal cord: 0.95 vs. 0.91). The proposed model was able to achieve higher detection accuracy compared with CNN model. As shown in figure 4, we showed some errors detected by our method and missed by the traditional CNN method.

| OAR | BA | | F score | | Sensitivity | | Specificity | | AUC | |
|---|---|---|---|---|---|---|---|---|---|---|
| | CNN | Proposed | CNN | Proposed | CNN | Proposed | CNN | Proposed | CNN | Proposed |
| Esophagus | 0.92 | 0.96 | 0.93 | 0.97 | 0.93 | 0.96 | 0.91 | 0.98 | 0.95 | 0.96 |
| Heart | 0.95 | 0.98 | 0.96 | 0.98 | 0.96 | 0.97 | 0.94 | 1.00 | 0.95 | 0.97 |
| Left lung | 0.98 | 0.99 | 0.91 | 0.98 | 0.96 | 0.98 | 1.00 | 1.00 | 0.93 | 0.97 |
| Right lung | 0.97 | 0.99 | 0.92 | 0.98 | 0.97 | 0.98 | 1.00 | 1.00 | 0.94 | 0.97 |
| Spinal cord | 0.91 | 0.96 | 0.92 | 0.96 | 0.93 | 0.94 | 0.89 | 0.98 | 0.91 | 0.95 |

**Table 1**. The comparison of CNN model and the proposed model on the test dataset.

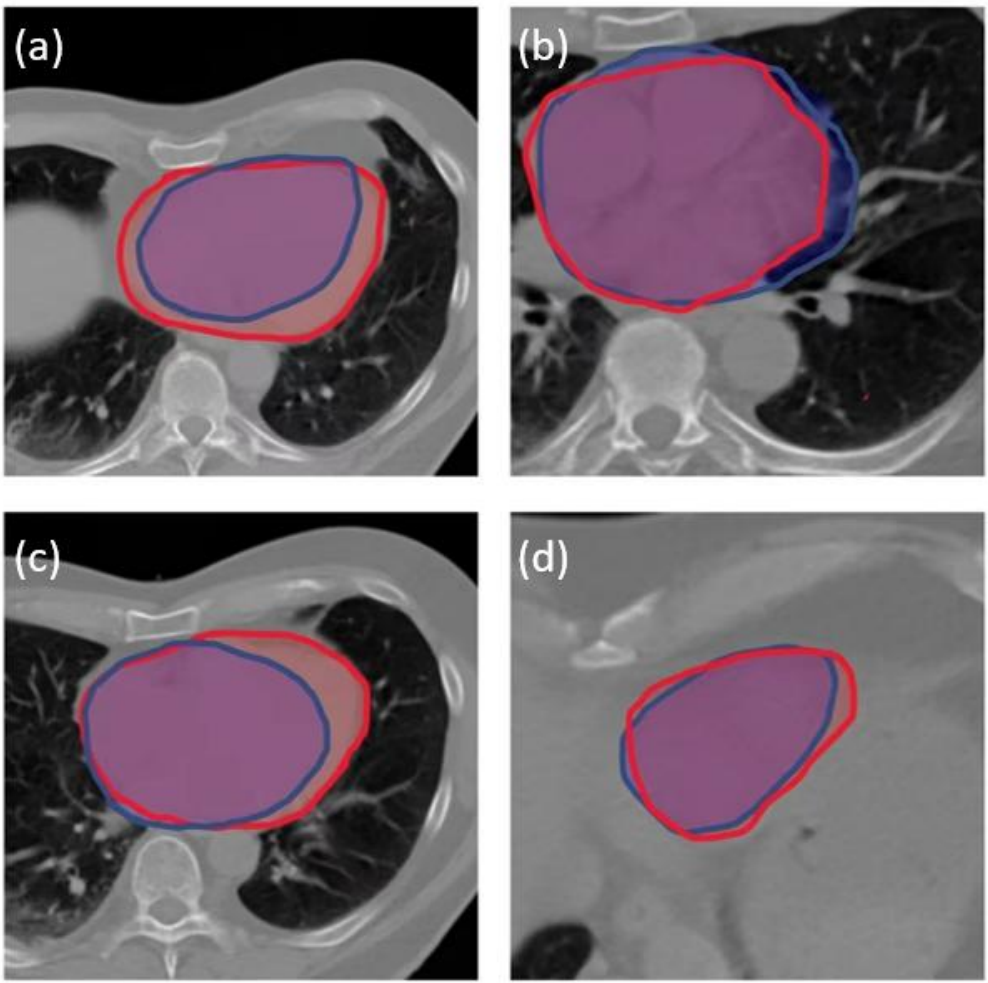

**Figure 4.** The comparation of the errors detected by our method and CNN method. The contour errors in (a) and (c) were detected by both CNN method and the proposed method. The contour errors in (b) and (d) were only detected only by the proposed method. The red line was gold standard. The blue line indicated the contour generated by deep learning technique.

### 3.2. The performance of the proposed method on generated dataset

The balanced accuracy of the predication on different generated errors was shown in Table 2. The proposed method achieved a higher balanced accuracy in identifying reduction errors compared to the CNN method (Esophagus: 0.82 vs 0.46; heart: 0.83 vs 0.67; left lung: 0.85 vs 0.66; right lung: 0.85 vs 0.66; spinal cord: 0.81 vs 0.44). In terms of identifying enlargement errors, the proposed method demonstrated a superior balanced accuracy in comparison to the CNN method (Esophagus: 0.88 vs 0.50; heart: 0.86 vs 0.69; left lung: 0.88 vs 0.70; right lung: 0.88 vs 0.70; spinal cord: 0.83 vs 0.49). The balanced accuracy of the proposed method in identifying translation errors surpassed that of the CNN method (Esophagus: 0.88 vs 0.51; heart: 0.89 vs 0.70; left lung: 0.89 vs 0.70; right lung: 0.89 vs 0.70; spinal cord: 0.87 vs 0.49). As shown in figure 5, the generated errors were detected by our method and missed by the traditional CNN method (Figure 5c & Figure 5d).

|  | Reduction | | Enlargement | | Translation | |
| --- | --- | --- | --- | --- | --- | --- |
|  | CNN | Proposed | CNN | Proposed | CNN | Proposed |
| Esophagus | 0.46 | 0.82 | 0.50 | 0.88 | 0.51 | 0.88 |
| Heart | 0.67 | 0.83 | 0.69 | 0.86 | 0.70 | 0.89 |
| Left lung | 0.66 | 0.85 | 0.70 | 0.88 | 0.70 | 0.89 |
| Right lung | 0.66 | 0.85 | 0.70 | 0.88 | 0.70 | 0.89 |
| Spinal cord | 0.44 | 0.81 | 0.49 | 0.83 | 0.49 | 0.87 |

Table 2. The comparison of the identification performance on generated dataset

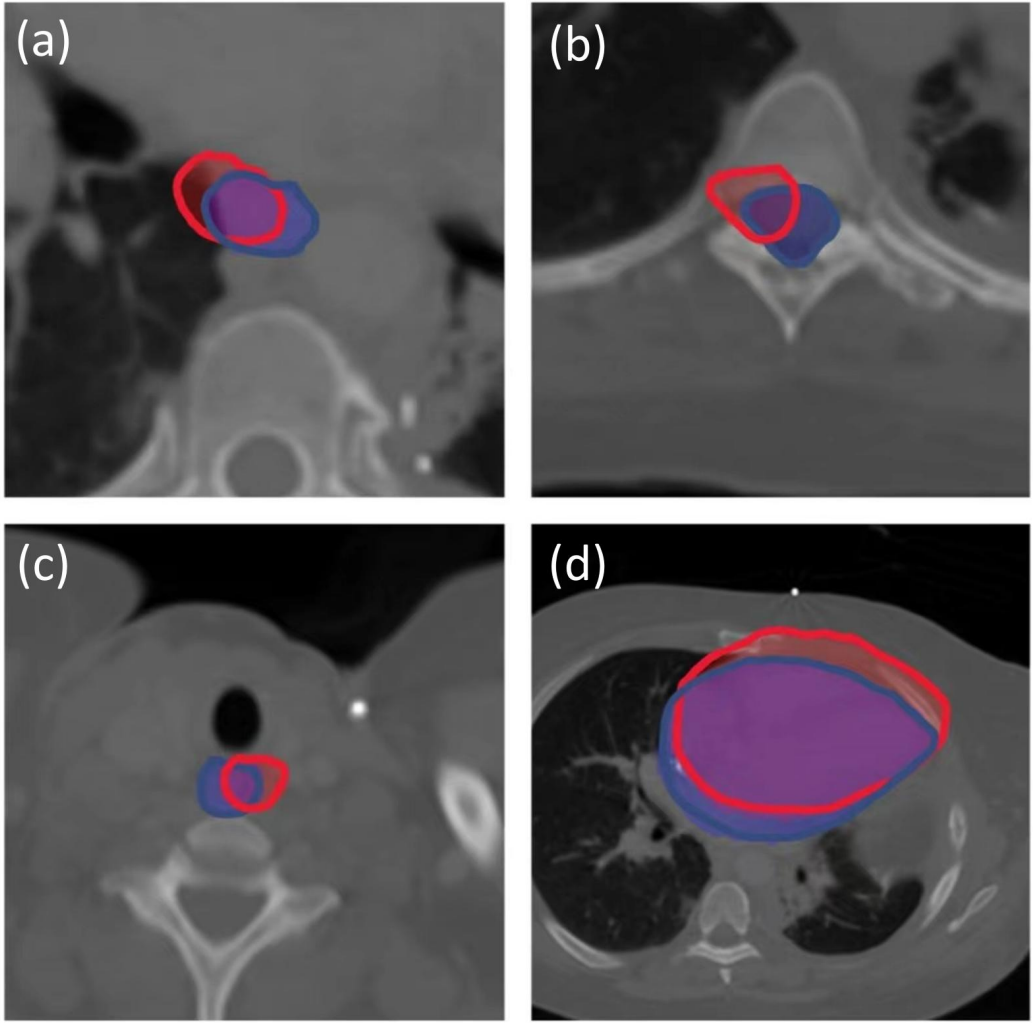

**Figure 5.** The comparation of the generated errors detected by our method and CNN method. The contour errors in (a) and (b) were detected by both CNN method and the proposed method. The contour errors in (c) and (d) were only detected only by the proposed method. The blue line was gold standard. The red line indicated the contour with generated errors.

### 3.3. Detection limit

The detection limits of the proposed method and CNN on generated dataset were shown in Table 3. The proposed method achieved higher detection limits than the CNN method (Esophagus: 3 vs 20; heart: 8 vs 24; left lung: 12 vs 35; right lung: 12 vs 34; spinal cord: 3 vs 18). To explore the factor associated with the detection limits, we computed the volume of each OAR (Table 3) and determined the Pearson correlation coefficient (r=0.94). To validate the detection limit, we evaluated the quality of the generated contours using metrics such as DSC, HD95, and MSD (Table 4). Notably, the detection limit exhibited a strong correlation with DSC (r=0.86), HD95 (r=0.97), and MSD (r=0.99).

| OARs | Esophagus | Heart | Left lung | Right lung | Spinal cord |
|---|---|---|---|---|---|
| CNN (unit: pixel) | 20 | 24 | 35 | 34 | 18 |
| Proposed (unit: pixel) | 3 | 8 | 12 | 12 | 3 |
| Volume (unit: pixel) | 17119 | 255252 | 421346 | 653508 | 23087 |
| DSC (mean±std) | 0.77±0.03 | 0.93±0.03 | 0.93±0.04 | 0.94±0.04 | 0.67±0.06 |
| HD$_{95}$ (mean±std) | 3.6±1.9 | 6.6 ± 3.3 | 11.1 ± 8.0 | 12.3 ± 8.0 | 2.98±3.45 |
| MSD (mean±std) | 0.50±0.35 | 1.43±0.65 | 2.25±0.55 | 2.30±0.55 | 0.59±0.34 |

Table 3. The detection limit and the related factor

## 4. Discussion

In this study, we developed an automatic QA model to detect various types of errors in the contours of OARs. This model is particularly beneficial for automatic contouring methods and less experienced physicians in standard delineation. To enable the model to recognize different types of errors, we proposed a one-class classifier for anomalous OAR detection. To enhance the recognition capabilities of the model, we proposed a novel method of training with cropped regions of interest. We evaluated the method using different metrics. The results demonstrated the strong performance and compatibility of the proposed method in recognizing various types of errors. Specifically, we confirmed the following findings: The proposed one-class classifier performed well in identifying various errors in OAR delineations, exhibiting strong generalizability. The addition of an attention mechanism effectively improved the recognition of erroneous samples by efficiently filtering out irrelevant information in the images. We introduced a method to dynamically assess the quality of a segmentation from multiple perspectives, accommodating the characteristics of various OARs. The method has low hardware requirements and runs quickly, making it easy to deploy on various computers.

Traditional approaches of contour error detection typically employed binary classifiers [24, 26]. In contrast, our study introduces a one-class classifier and attention mechanisms, which provided the model with increased versatility and enhanced precision compared to previous research. Binary classifiers may suffer from decreased performance when the test set contains errors that are significantly different from those in the training set. By leveraging the characteristics of a one-class classifier, our proposed method can enhance the detection capability for errors that have not been encountered before. The use of a one-class classifier effectively addresses the challenge of limited error samples, an issue that has been largely overlooked in previous studies [21,25,26]. The scarcity of error samples has hindered the performance of previous methods. However, even a simple approach of generating error samples, such as the method proposed in this paper, and training a binary classifier can significantly enhance performance. In Henderson's study[29], a CNN and a graph neural network (GNN) were combined to leverage the segmentation's appearance and shape, which automatically identified errors in 3D OAR segmentations in CT scans. To create the training and test dataset, they perturbed each ground-truth segmentation of the parotid gland by 100 times [29]. The implementation of their method in three dimensions required 3D kernels of convolution, which might lead to high computation demands for the process. To mitigate the issue of imbalanced training samples, a weighted cross-entropy loss function or cross-validation method can also be used in model training.

To validate the performance of our model in clinical scenarios, we sought to obtain results that would effectively demonstrate its capabilities. In this context, we investigated the model's detection limit. Our analysis revealed a consistent phenomenon: the volume, DSC, HD$_{95}$, and MSD of OARs exhibited a proportional relationship with their respective detection limits. This observation suggests that for larger OARs, errors in their delineation may be sufficiently substantial to impact the outcomes of radiotherapy. In the future it may be necessary to establish uniform low-quality standards for different OARs. [26] However, given that automatic delineation algorithms may exhibit varying accuracy in delineating different low-quality OARs, this approach could potentially misclassify the delineation of certain organs as low quality. To address this challenge, our future work would involve considering clinical treatment requirements and establishing organ-specific criteria for poor-quality delineation.

Numerous previous studies[21,23,25,26] have demonstrated the successful application of these quality control methods to various disease cites. However, it is important to note that retraining the model may be necessary for each specific organ at risk (OAR). Additionally, challenges may arise due to unclear boundaries between certain organs and surrounding soft tissues, potentially leading to delineation errors. To address this issue, the optimization of window width and window level of the CT images may be required. Hang et al. proposed a method that utilized rigid registration to align the delineations of computed tomography (CT) images with those of magnetic resonance imaging (MRI) images. [44] Various neural networks were employed to delineate MRI images. The results demonstrated the effectiveness of their method, supporting the compatibility of deep learning-based approaches across different modalities of images. Based on these findings, we could infer that the proposed model has the potential to be extended to other modalities of medical images, such as MRI, PET-CT, and so on.

This study has several limitations. Firstly, the employment of the memory-intensive ResNet-152 network may present challenges in deployment on low-performance computers with limited GPU memory. Secondly, the dataset used for training and evaluation is relatively homogeneous, potentially introducing biases when the network is applied to datasets from other healthcare institutions, as different physicians may have varying segmentation preferences. Lastly, the dynamic range of computed tomography (CT) images is a critical factor influencing model performance. In this study, CT images were normalized to 8 bits, while clinical CT images are typically 16 bits, which may result in the loss of some image information.

Future research directions include the development of lighter networks, such as MobileNet, to achieve similar performance while enhancing the method's versatility and applicability on devices with limited computational resources. We also plan to develop new network architectures specifically tailored to the characteristics of medical images. To improve the generalizability and accuracy of the model, we encourage more medical institutions to share their datasets. This will facilitate the development of more robust and reliable models. Furthermore, we envision the integration of validated large-scale models into our research, leveraging their robust compatibility and potential for recognizing various types of errors. Additionally, we aim to contribute to the establishment of industry-wide standards for deploying this research in the cloud. This will simplify the adoption and utilization of the method by healthcare institutions, potentially benefiting a larger patient population. With access to larger and higher-quality datasets, we will be able to develop increasingly accurate and effective models.

5. Conclusions

In this study, we present a novel quality assurance (QA) model for organ-at-risk (OAR) delineation in radiotherapy. This model incorporates two key innovations: a one-class classifier and an attention mechanism. These advancements enhance the model's generalization capabilities and accuracy in detecting various types of delineation errors. The model's ability to detect a wide range of errors makes it a valuable tool for ensuring the quality and consistency of OAR delineations. Furthermore, the model's rapid execution speed significantly reduces the time and effort required for physicians to perform QA tasks, thereby streamlining the radiotherapy planning process.

**References**

1. Kong, F.J.C.; cancer, m.t.i.l.a.n.-s.c.l. Randomized phase II trial of individualized adaptive radiotherapy using during-treatment FDG-PET. **2016**.
2. Liao, Z.; Bradley, J.; Choi, N.J.S.a. RTOG 1308: Phase III randomized trial comparing overall survival after photon versus proton chemotherapy for inoperable stage II-IIIB NSCLC. 2015https.
3. Ritter, T.; Quint, D.J.; Senan, S.; Gaspar, L.E.; Komaki, R.U.; Hurkmans, C.W.; Timmerman, R.; Bezjak, A.; Bradley, J.D.; Movsas, B.J.I.J.o.R.O.B.P. Consideration of dose limits for organs at risk of thoracic radiotherapy: atlas for lung, proximal bronchial tree, esophagus, spinal cord, ribs, and brachial plexus. **2011**, *81*, 1442-1457.
4. Mir, R.; Kelly, S.M.; Xiao, Y.; Moore, A.; Clark, C.H.; Clementel, E.; Corning, C.; Ebert, M.; Hoskin, P.; Hurkmans, C.W.J.R.; et al. Organ at risk delineation for radiation therapy clinical trials: Global Harmonization Group consensus guidelines. **2020**, *150*, 30-39.
5. Cabezas, M.; Oliver, A.; Lladó, X.; Freixenet, J.; Cuadra, M.B.J.C.m.; biomedicine, p.i. A review of atlas-based segmentation for magnetic resonance brain images. **2011**, *104*, e158-e177.

6. Bach Cuadra, M.; Duay, V.; Thiran, J.-P.J.H.o.B.I.M.; Research, C. Atlas-based segmentation. **2015**, 221-244.
7. Ronnerberger, O.; Fischer, P.; Brox, T. U-Net: Convolutional Neural Networks for Biomedical Image Segmentation. In Proceedings of the Medical Image Computing and Computer-Assisted Intervention—MICCAI, 2015.
8. Çiçek, Ö.; Abdulkadir, A.; Lienkamp, S.S.; Brox, T.; Ronneberger, O. 3D U-Net: learning dense volumetric segmentation from sparse annotation. In Proceedings of the Medical Image Computing and Computer-Assisted Intervention–MICCAI 2016: 19th International Conference, Athens, Greece, October 17-21, 2016, Proceedings, Part II 19, 2016; pp. 424-432.
9. Milletari, F.; Navab, N.; Ahmadi, S.-A. V-net: Fully convolutional neural networks for volumetric medical image segmentation. In Proceedings of the 2016 fourth international conference on 3D vision (3DV), 2016; pp. 565-571.
10. Badrinarayanan, V.; Kendall, A.; Cipolla, R.J.I.t.o.p.a.; intelligence, m. Segnet: A deep convolutional encoder-decoder architecture for image segmentation. **2017**, *39*, 2481-2495.
11. Kamnitsas, K.; Ledig, C.; Newcombe, V.F.; Simpson, J.P.; Kane, A.D.; Menon, D.K.; Rueckert, D.; Glocker, B.J.M.i.a. Efficient multi-scale 3D CNN with fully connected CRF for accurate brain lesion segmentation. **2017**, *36*, 61-78.
12. Chen, L.-C.; Papandreou, G.; Kokkinos, I.; Murphy, K.; Yuille, A.L.J.I.t.o.p.a.; intelligence, m. Deeplab: Semantic image segmentation with deep convolutional nets, atrous convolution, and fully connected crfs. **2017**, *40*, 834-848.
13. Chen, H.; Dou, Q.; Yu, L.; Qin, J.; Heng, P.-A.J.N. VoxResNet: Deep voxelwise residual networks for brain segmentation from 3D MR images. **2018**, *170*, 446-455.
14. He, K.; Gkioxari, G.; Dollár, P.; Girshick, R. Mask r-cnn. In Proceedings of the Proceedings of the IEEE international conference on computer vision, 2017; pp. 2961-2969.
15. Fidon, L.J.a.p.a. Trustworthy Deep Learning for Medical Image Segmentation. **2023**.
16. Matoska, T.; Patel, M.; Liu, H.; Beriwal, S.J.A.i.R.O. Review of Deep Learning Based Autosegmentation for Clinical Target Volume–Current Status and Future Directions. **2024**, 101470.
17. Taha, A.A.; Hanbury, A.J.B.m.i. Metrics for evaluating 3D medical image segmentation: analysis, selection, and tool. **2015**, *15*, 1-28.
18. Wang, T.; Chen, Y.; Qiao, M.; Snoussi, H.J.T.I.J.o.A.M.T. A fast and robust convolutional neural network-based defect detection model in product quality control. **2018**, *94*, 3465-3471.
19. Rhee, D.J.; Akinfenwa, C.P.A.; Rigaud, B.; Jhingran, A.; Cardenas, C.E.; Zhang, L.; Prajapati, S.; Kry, S.F.; Brock, K.K.; Beadle, B.M.J.J.o.a.c.m.p. Automatic contouring QA method using a deep learning–based autocontouring system. **2022**, *23*, e13647.
20. Nourzadeh, H.; Hui, C.; Ahmad, M.; Sadeghzadehyazdi, N.; Watkins, W.T.; Dutta, S.W.; Alonso, C.E.; Trifiletti, D.M.; Siebers, J.V.J.M.p. Knowledge-based quality control of organ delineations in radiation therapy. **2022**, *49*, 1368-1381.
21. Men, K.; Geng, H.; Biswas, T.; Liao, Z.; Xiao, Y.J.F.i.O. Automated quality assurance of OAR contouring for lung cancer based on segmentation with deep active learning. **2020**, *10*, 986.
22. McIntosh, C.; Svistoun, I.; Purdie, T.G.J.I.t.o.m.i. Groupwise conditional random forests for automatic shape classification and contour quality assessment in radiotherapy planning. **2013**, *32*, 1043-1057.
23. Kis, A.; Kovács, F.; Szolgay, P. Analogic CNN algorithms for textile quality control based on optical and tactile sensory inputs. In Proceedings of the Proc. of The 8th IEEE International Biannual Workshop on Cellular Neural Networks and their Applications, Budapes, 2004.
24. Hui, C.B.; Nourzadeh, H.; Watkins, W.T.; Trifiletti, D.M.; Alonso, C.E.; Dutta, S.W.; Siebers, J.V.J.M.p. Quality assurance tool for organ at risk delineation in radiation therapy using a parametric statistical approach. **2018**, *45*, 2089-2096.
25. Duan, J.; Bernard, M.E.; Castle, J.R.; Feng, X.; Wang, C.; Kenamond, M.C.; Chen, Q.J.M.P. Contouring quality assurance methodology based on multiple geometric features against deep learning auto-segmentation. **2023**.
26. Chen, X.; Men, K.; Chen, B.; Tang, Y.; Zhang, T.; Wang, S.; Li, Y.; Dai, J.J.F.i.O. CNN-based quality assurance for automatic segmentation of breast cancer in radiotherapy. **2020**, *10*, 524.
27. Altman, M.; Kavanaugh, J.; Wooten, H.; Green, O.; DeWees, T.; Gay, H.; Thorstad, W.; Li, H.; Mutic, S.J.P.i.M.; Biology. A framework for automated contour quality assurance in radiation therapy including adaptive techniques. **2015**, *60*, 5199.

28. Zhang, Y.; Plautz, T.E.; Hao, Y.; Kinchen, C.; Li, X.A.J.M.p. Texture-based, automatic contour validation for online adaptive replanning: a feasibility study on abdominal organs. **2019**, *46*, 4010-4020.
29. Henderson, E.G.; Green, A.F.; van Herk, M.; Vasquez Osorio, E.M. Automatic identification of segmentation errors for radiotherapy using geometric learning. In Proceedings of the International Conference on Medical Image Computing and Computer-Assisted Intervention, 2022; pp. 319-329.
30. He, K.; Zhang, X.; Ren, S.; Sun, J. Deep residual learning for image recognition. In Proceedings of the Proceedings of the IEEE conference on computer vision and pattern recognition, 2016; pp. 770-778.
31. Long, J.; Shelhamer, E.; Darrell, T. Fully convolutional networks for semantic segmentation. In Proceedings of the Proceedings of the IEEE conference on computer vision and pattern recognition, 2015; pp. 3431-3440.
32. Uijlings, J.R.; Van De Sande, K.E.; Gevers, T.; Smeulders, A.W.J.I.j.o.c.v. Selective search for object recognition. **2013**, *104*, 154-171.
33. Khan, S.S.; Madden, M.G. A survey of recent trends in one class classification. In Proceedings of the Artificial Intelligence and Cognitive Science: 20th Irish Conference, AICS 2009, Dublin, Ireland, August 19-21, 2009, Revised Selected Papers 20, 2010; pp. 188-197.
34. Tax, D.M.J. One-class classification: Concept learning in the absence of counter-examples. **2002**.
35. Schölkopf, B.; Platt, J.C.; Shawe-Taylor, J.; Smola, A.J.; Williamson, R.C.J.N.c. Estimating the support of a high-dimensional distribution. **2001**, *13*, 1443-1471.
36. Ruff, L.; Vandermeulen, R.; Goernitz, N.; Deecke, L.; Siddiqui, S.A.; Binder, A.; Müller, E.; Kloft, M. Deep one-class classification. In Proceedings of the International conference on machine learning, 2018; pp. 4393-4402.
37. Oza, P.; Patel, V.M.J.I.S.P.L. One-class convolutional neural network. **2018**, *26*, 277-281.
38. Yang, J.; Veeraraghavan, H.; Armato, S.G., III; Farahani, K.; Kirby, J.S.; Kalpathy-Kramer, J.; van Elmpt, W.; Dekker, A.; Han, X.; Feng, X.; et al. Autosegmentation for thoracic radiation treatment planning: A grand challenge at AAPM 2017. *Medical Physics* **2018**, *45*, 4568-4581, doi:10.1002/mp.13141.
39. Machtay, M.; Matuszak, M.; Bradley, J.; Hirsh, V.; Ten Haken, R.; Pryma, D.J.U. NRG ONCOLOGY ECOG-ACRIN RTOG 1106/ACRIN 6697 RANDOMIZED PHASE II TRIAL OF INDIVIDUALIZED ADAPTIVE RADIOTHERAPY USING DURING-TREATMENT FDG-PET/CT AND MODERN TECHNOLOGY IN LOCALLY. **2014**.
40. Sorensen, T.J.B.s. A method of establishing groups of equal amplitude in plant sociology based on similarity of species content and its application to analyses of the vegetation on Danish commons. **1948**, *5*, 1-34.
41. Blumberg, H. Hausdorff's Grundzüge der Mengenlehre. **1920**.
42. Heimann, T.; Van Ginneken, B.; Styner, M.A.; Arzhaeva, Y.; Aurich, V.; Bauer, C.; Beck, A.; Becker, C.; Beichel, R.; Bekes, G.J.I.t.o.m.i. Comparison and evaluation of methods for liver segmentation from CT datasets. **2009**, *28*, 1251-1265.
43. Men, K.; Boimel, P.; Janopaul-Naylor, J.; Zhong, H.; Huang, M.; Geng, H.; Cheng, C.; Fan, Y.; Plastaras, J.P.; Ben-Josef, E.J.P.i.M.; et al. Cascaded atrous convolution and spatial pyramid pooling for more accurate tumor target segmentation for rectal cancer radiotherapy. **2018**, *63*, 185016.
44. Min, H.; Dowling, J.; Jameson, M.G.; Cloak, K.; Faustino, J.; Sidhom, M.; Martin, J.; Ebert, M.A.; Haworth, A.; Chlap, P.J.P.i.M.; et al. Automatic radiotherapy delineation quality assurance on prostate MRI with deep learning in a multicentre clinical trial. **2021**, *66*, 195008.